\documentclass[manuscript,screen]{acmart}

\AtBeginDocument{%
  \providecommand\BibTeX{{%
    \normalfont B\kern-0.5em{\scshape i\kern-0.25em b}\kern-0.8em\TeX}}}

\setcopyright{acmcopyright}
\copyrightyear{2022}
\acmYear{2022}
\acmDOI{XXXXXXX.XXXXXXX}

\usepackage{latexsym}
\usepackage{xcolor,colortbl}

\usepackage[T1]{fontenc}
\usepackage[utf8]{inputenc}
\usepackage{todonotes}

\usepackage{color}

\usepackage{microtype}
\usepackage{array,multirow,graphicx}
\usepackage{float}
\usepackage{enumitem}

\usepackage{caption}
\usepackage{subcaption}
\begin{document}

\title{Impact of Tokenization on Language Models: An Analysis for Turkish}


\author{Cagri Toraman, Eyup Halit Yilmaz, Furkan \c{S}ah\.{i}nu\c{c}, \texorpdfstring{\MakeLowercase{and}}{and} Oguzhan Ozcelik}
\affiliation{%
  \institution{\\ Aselsan Research Center}
  \city{Ankara}
  \country{Turkey}}
\email{emails: {ctoraman, ehyilmaz, fsahinuc, ogozcelik}@aselsan.com.tr}




\renewcommand{\shortauthors}{Toraman et al.}

\begin{abstract}
Tokenization is an important text preprocessing step to prepare input tokens for deep language models. WordPiece and BPE are de facto methods employed by important models, such as BERT and GPT. However, the impact of tokenization can be different for morphologically rich languages, such as Turkic languages, where many words can be generated by adding prefixes and suffixes. We compare five tokenizers at different granularity levels, i.e. their outputs vary from smallest pieces of characters to the surface form of words, including a Morphological-level tokenizer. We train these tokenizers and pretrain medium-sized language models using RoBERTa pretraining procedure on the Turkish split of the OSCAR corpus. We then fine-tune our models on six downstream tasks. Our experiments, supported by statistical tests, reveal that Morphological-level tokenizer has challenging performance with de facto tokenizers. Furthermore, we find that increasing the vocabulary size improves the performance of Morphological and Word-level tokenizers more than that of de facto tokenizers. The ratio of the number of vocabulary parameters to the total number of model parameters can be empirically chosen as 20\% for de facto tokenizers and 40\% for other tokenizers to obtain a reasonable trade-off between model size and performance.
\end{abstract}



\maketitle

\section{Introduction}
Deep language models gained popularity with the introduction of masked language modeling based on the Transformer architecture \cite{Vaswani:2017} to pretrain a general purpose language understanding with BERT \cite{Devlin:2019} and its variants. The language models are then able to transfer the pretrained knowledge to downstream tasks, such as Sentiment Analysis and Named Entity Recognition. Indeed, such large models provide impressing results on the performance of many downstream tasks, not only in natural language processing \cite{Devlin:2019}, but also many other research areas such as search \cite{Yates:2021} and recommendation \cite{Sun:2019}.

Tokenization is an important text preprocessing step for deep language models. Conventional word embeddings, such as word2vec \cite{Mikolov:2013}, generally use vocabularies consisting of the surface forms of words. On the other hand, deep language models employ more efficient tokenization algorithms where input text is split into smaller pieces so that out-of-vocabulary words can still be processed. Language models can also benefit from the tokens that represent basic semantic units to better comprehend text semantics.

Transformer-based language models generally employ two de facto tokenization algorithms, namely WordPiece \cite{Schuster:2012} and Byte Pair Encoding (BPE) \cite{Sennrich:2016}. For instance, BERT \cite{Devlin:2019} uses WordPiece, whereas GPT-2 \cite{Radford:2019} employs BPE. However, large language models are first pretrained for English; successor pretrained models in low-resource languages thereby employ the same tokenizers \cite{Schweter:2020}. The impact of tokenization algorithms can be different for low-resource languages, such as agglutinative Turkic and Uralic languages, where words can have prefixes and suffixes. For instance, in Turkish, parsing the word "veremedim" (translated as "I could not give") results in "ver-e-me-di-m" including four suffixes in a single word. A Morphological-level tokenizer can output five tokens in this case, providing the model with a better understanding of word semantics. An example benefit is that the language model would relate the suffix "-me" to negation, similar to the word "not" in English. Moreover, the impact of different tokenization methods, including token representation in different levels from Character-level to Word-level, is not examined in details for low-resource languages, specifically for Turkish. 

The number of unique tokens used in training deep language models is referred to as the vocabulary size. Any evaluation data would be split into tokens by using the selected tokenization algorithm according to the trained vocabulary. There is a likelihood of observing out-of-vocabulary or unknown tokens, i.e. some tokens in the evaluation data can be missing in the trained vocabulary. The problem with unknown tokens is that they are mapped to the same embedding without semantic context, resulting in possible performance loss. As the vocabulary size increases, the likelihood of getting such unknown tokens is decreased, since the vocabulary would capture more instances of tokens. On the other hand, the model becomes less efficient in terms of its size, i.e, the memory requirement would increase and the model would become more costly to train. This results in a trade-off between model size and performance in terms of vocabulary size. 

In this study, we thereby examine the following research questions.
\begin{itemize}
    \item \textbf{RQ-1}. What is the impact of different tokenization methods on the performance of Turkish language modeling and varying downstream tasks? e.g. Does a Morphological-level tokenization method have benefits on Turkish language modeling?
    \item \textbf{RQ-2}. How does the model performance change in different tokenization methods when the vocabulary size is tuned for the trade-off between model size and performance?
\end{itemize}

In order to answer our research questions, we compare the performance of different tokenization methods for Turkish. We select five tokenizers at different granularity levels, i.e. their outputs vary from smallest pieces (characters) to the surface form (words), which are Character-level, BPE, WordPiece, Morphological-level, and Word-level tokenization, respectively. In order to evaluate their performances, we train a tokenizer for each method, and pretrain medium language models using RoBERTa \cite{Liu:2019} pretraining procedure on the Turkish split of the OSCAR \cite{Suarez:2019} corpus, called \textbf{RoBERTa-TR-medium}\footnote{We publish our pretrained models with different tokenizers and vocabulary sizes at https://huggingface.co/ctoraman}. We then evaluate the performance of our models by fine-tuning them on six downstream tasks; namely News Classification, Hate Speech Detection, Sentiment Analysis, Named Entity Recognition, Semantic Text Similarity, and Natural Language Inference. 

The \textbf{main contributions and practical implications} of this study can be summarized as follows.
\begin{itemize}
    \item We analyze the impact of tokenizers, at different granularity levels from character to word-level, on varying downstream tasks for Turkish language models. We find that Morphological-level tokenizer is competitive with de facto tokenizers, i.e. BPE and WordPiece. Our experimental results, supported by statistical tests, can shed light on the role of tokenization in language modeling, specifically for morphologically rich languages.
    
    \item We show that increasing the vocabulary size improves the performances of Morphological and Word-level tokenizers more than that of de facto tokenizers, BPE and WordPiece. The ratio of the number of vocabulary parameters to the total number of model parameters can be empirically chosen as 20\% for de facto tokenizers and 40\% for others. This choice can result in a more efficient use of computational resources, where the majority of the parameters are allocated to the Transformer blocks instead of the vocabulary embeddings. Moreover, we believe that our experimental results on the vocabulary size can provide an empirical guidance for other researchers who work on pretraining deep language models.
    
    \item We compare our medium-based models with a state-of-the-art Turkish language model \cite{Schweter:2020} that has the same model architecture with BERT-base \cite{Devlin:2019}, and show that our approximately 3-times smaller model can recover 97\% of the performance of the larger one. Our language models are publicly available, so that other researchers and practitioners can benefit from our models in terms of a more efficient model size and varying vocabulary sizes. This would reduce memory requirements, and also provide responsible energy usage and smaller carbon footprint in return, which we discuss in Section \ref{section:discussion} in details, along with other ethical concerns including transparency and fairness.
\end{itemize}

The rest of the paper is organized as follows. In the next section, we provide a brief literature review of tokenization algorithms in general and also tokenization in low-resource languages. We explain the details of tokenization methods, our pretrained model, and downstream tasks in Section \ref{section:impact}. Our comparative experiments on different tokenizers and analysis on vocabulary size are given in Section \ref{section:experiments}. We then provide a short discussion on the ethical concerns and broader impact of our study in Section \ref{section:discussion}. We conclude the study in the last section.

\section{Related Work}

\subsection{Tokenization Algorithms}

Since tokenization is one of the first steps in any Information Retrieval or Natural Language Processing system, the importance of using a tokenization algorithm is highlighted in early studies \citep{Jimenez:2011}. The prevalent tokenization algorithms in the literature, Byte Pair Encoding (BPE) \cite{Sennrich:2016} and WordPiece \cite{Schuster:2012}, are of recent interest in language model pretraining research. Many noteworthy studies in the literature focus on enhancing these subword tokenization methods. For example, \citet{Ding:2019} explore the impact of the number of BPE merges on the machine translation performance. \citet{Provilkov:2020} propose a drop-out method for each merge step of BPE in order to break the deterministic nature of BPE, which provides a performance improvement in machine translation.

BPE is found to be suboptimal for language pretraining \cite{Bostrom:2020} as it does not effectively utilize the vocabulary space. \citet{Nayak:2020} compare the activations of attention layers of BERT with WordPiece and Word-level tokenization to assess the effect of including subword tokens. They find out that the vocabulary with frequency-based character combinations hinders the ability of modeling semantically meaningful relations between words. Additionally, tokenization based on word occurrence statistics results in representations that are dependent on frequency information rather than semantics \cite{Gong:2018}. On the other hand, it is proposed to apply subword regularization by utilizing multiple subword segmentations to enhance the robustness of the neural machine translation models \citep{Kudo:2018a}. Based on this algorithm, which implements BPE and Unigram language models, SentencePiece is proposed as another tokenization method \citep{Kudo:2018b}. Recently, \citet{Xu:2021} approach the problem of finding the best token vocabulary with a proper size in the scope of the trade-off between vocabulary entropy and vocabulary size. The produced vocabularies in diverse scenarios achieve both reduced sizes and performance improvements. In addition, learning optimal vocabulary takes significantly less time than regular BPE-search approach.

Alternative tokenization algorithms using morphological analysis can be promising candidates for subword tokenization that increase training efficiency and downstream performance \cite{Park:2020}. Rule-based tokenization algorithms utilizing lexicons and semantic parsing can extend existing methods to cross-lingual settings \cite{Vasiu:2020}. Joint and hybrid tokenization approaches combine coarse and fine-grained representations to incorporate Word-level and subword representations \cite{Hiraoka:2021}. Multi-grained tokenization methods are incorporated into the model architecture to capture multi-word representations, such as \emph{ice cream}, at the expense of increased computational complexity \cite{Xinsong:2021}. Enabling a gradient-based learnable representation in the tokenization step of the pipeline is an emerging line of research \cite{Tay:2021}. In our study, we provide a comprehensive analysis on the impact of tokenization algorithms at different granularity levels from character to word-level, and evaluate the performances on a diverse range of downstream tasks.

\subsection{Tokenization in Low-resource Languages}

Tokenization-based methods aiming to enhance the downstream task's performance in low-resource languages have been studied before introducing the de facto tokenization methods \citep{Kulick:2011}. With the emergence of the de facto tokenizers, the effects of SentencePiece, Word-level, and Syllable-level tokenization strategies are investigated for low-resource languages, such as Thai \cite{Lowphansirikul:2021}. In addition, \citet{Li:2021} show that character-based subword tokenization methods give better results than syllable-based ones in Tibetan-to-Chinese machine translation.

Part-of-Speech Tagging is one of the downstream tasks where different tokenization-based methods are employed in low resource languages \citep{Kaing:2021,Dingchen:2018,Dingchen:2019}. Morphological analysis is used to propose a tokenization system for Kurdish \cite{Ahmadi:2020}. Exploiting pretrained models with parameter freezing and additional intermediate layers is beneficial for Uyghur-Chinese machine translation \cite{Zhang:2021c}. \citet{Dossou:2021} propose a phrase-based tokenization method for neural machine translation task between Fon language and French. Since Fon language is quite specific and low-resource, bilingual people are involved in data cleaning and preprocessing phases to extract best phrases based on the linguistic components of the Fon language. \citet{Park:2021} studies morphological features of the Korean language while training a tokenizer in the scope of machine translation. During training tokenizers, target sentences that are not processed by morphological analysis, are also utilized. 

Although there are some efforts for pretraining Turkish language models \citep{TurkishLMs:2020,Schweter:2020}, the effect of tokenization algorithms including a Morphological-level one is yet to be studied. To the best of our knowledge, this is the first study that investigates the impact of tokenization and vocabulary size on Turkish.

\section{Impact of Tokenization}
\label{section:impact}
In order to understand the impact of different tokenization methods on language modeling, we first explain the tokenization approaches that are examined in this study. We then introduce our pipeline that describes the details of various steps to obtain language models with different tokenizers.

\subsection{Tokenization Methods}
\label{section:tokenization_methods}
In our study, we consider five tokenization algorithms making use of different linguistic features including characters, frequency, and grammatical rules, explained with respect to the granularity levels as follows.

\begin{itemize}[leftmargin=*]
    \item \textbf{Character-level}: Unlike the tokenization methods performing on word or subword units, Character-level tokenizers split words into the smallest parts. Since Character-level tokenizer requires no training to learn a vocabulary, we employ the ByT5 tokenization \cite{Xue:2021}. The advantage of this type of tokenization is that they can be utilized in any language to represent any character sequence in byte-level and enable a diverse modeling. Character-level tokenization also reduces the memory requirement in terms of model size, since it has a very limited number of tokens in the vocabulary. A disadvantage of this approach is that the model has to spend more capacity to reach at a higher-level representation compared to other tokenization methods. For instance, the language model has to learn during training that "t" and "h" co-occur frequently in English, whereas another tokenizer can provide this information to the models with a "th" token. Furthermore, the output for a given sequence would contain a large number of tokens when compared to other tokenizers. This results in potential information loss, since deep language models have an input parameter of text sequence length. 
    
    \item \textbf{BPE}: Byte Pair Encoding (BPE) is a frequently used tokenizer for pretrained language models \cite{Sennrich:2016}. The granularity of BPE can be considered as mid-level between character and word-level, such that tokens are mostly subwords depending on vocabulary size. In this method, all unique words are first extracted. A base vocabulary is then constructed from all symbols occurring in the unique words. The final vocabulary is built by merging the symbols according to the frequencies of consecutive symbols or subwords. Since BPE operates with byte representations, the vocabulary can encompass tokens from multiple languages and informal character sequences such as emojis. 
    
    \item \textbf{WordPiece}: Similar to BPE, WordPiece is also based on merging characters in the documents \cite{Schuster:2012}. Its main difference from BPE is that, WordPiece merges symbols towards maximizing a likelihood score of language modeling, i.e., when the probability of the merged symbol divided by individual probabilities of the symbols is greater than any other symbol pair. WordPiece and BPE are frequency-based algorithms that aim to increase the modeling power of individual tokens while being able to tokenize words that are not encountered during training of the tokenizer.
    
    \item \textbf{Morphological-level}: Morphological analysis can provide suffixes and word stems that are semantically more meaningful and valuable than the tokens obtained with overlapping frequency or likelihood. We therefore examine using the parsing output (without tags) of morphological analysis as input tokens. We use Zemberek morphological analysis tool for Turkish \cite{Akin:2007} before training the tokenizer. The advantage of Morphological-level tokenization is to capture grammatically interpretable character sequences in modeling and learn the semantics based on the suffixes of words. A disadvantage of this approach is that word stems are not split further and constitute a large set that has to be included in the vocabulary.
    
    \item \textbf{Word-level}: The granularity of Word-level tokenizer is surface forms of words, i.e. splits text according to the spaces between words. Word-level tokenization requires no vocabulary training, since one can apply it by just splitting text with white space characters. One explicit disadvantage is that this tokenizer requires more vocabulary size to properly tokenize the same amount of text compared to other methods. Since vocabulary has a limited size in language modeling, out-of-vocabulary or unknown tokens are likely to be observed in this approach.
    
\end{itemize}

The sample outputs provided by different tokenization methods are given for a sample sentence "Toplumsal barış sağlanır" (translated as "Social peace would be achieved") in Table \ref{tab:tokenization_examples}. All tokenizers have a vocabulary size of 16.6k tokens in this example, except that Character-level tokenizer has a vocabulary size of 384 characters. We note that BPE and WordPiece tokenizers can assign the surface forms of words to tokens, while Word-level tokenizer fails to capture some words and produces unknown tokens. The reason could be that the vocabulary capacity is utilized more efficiently by BPE and WordPiece tokenizers, whereas Word-level tokenizer fills up the vocabulary with more frequent words and cannot tokenize less frequent words. Morphological-level tokenizer overcomes this by assigning individual tokens to suffixes and isolating the word stems. We note that the output sequence length of Character-level tokenization is considerably higher than other tokenizers, which is not practical when language model requires a limited length of input text sequence (e.g. if this length parameter is set to 10 tokens, then all methods can properly represent the input, except that Character-level tokenizer can only represent its first 10 tokens or characters).

\begin{table}[t]
    \centering
    \begin{tabular}{|l|l|}
    \hline
         \textbf{Method} & \textbf{Tokenized text} \\
    \hline
        Character-level & "t", "o", "p", "l", "u", "m", "s", "a", "l", " ", "b", "a", "r", "ı", "ş", " ", "s", "a", "ğ", "l", "a", "n", "ı", "r" \\
        BPE & "[CLS]", "toplumsal", "barış", "sağ", "\#\#lanır", "[SEP]"\\
        WordPiece & "[CLS]", "toplumsal", "barış", "sağlan", "\#\#ır", "[SEP]" \\
        Morphological-level & "[CLS]", "toplum", "\#\#sal", "barış", "sağ", "\#\#lanır", "[SEP]" \\
        Word-level & "[CLS]", "[UNK]", "barış", "[UNK]", "[SEP]" \\
    \hline
    \end{tabular}
    \caption{The outputs of different tokenization methods for a sample input, "Toplumsal barış sağlanır" (translated as "Social peace would be achieved").
    }
    \label{tab:tokenization_examples}
\end{table}

\subsection{Our Pretrained Model: RoBERTa-Turkish-medium}
\label{sec:our_model}
\begin{figure}
    \centering
    \includegraphics[width=\columnwidth]{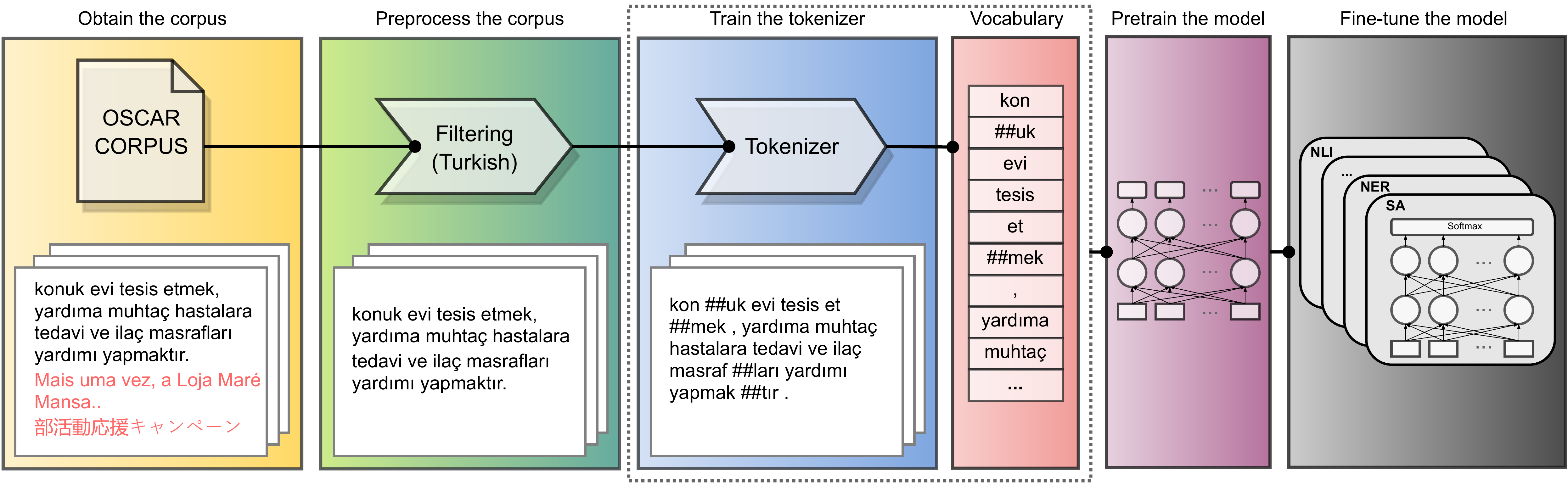}
    \caption{An illustration of our pretraining pipeline. There are five steps of pretraining process. We first obtain a text corpus (\textit{OSCAR Turkish deduplicated}), and preprocess the corpus by filtering non-Turkish texts. We then choose a tokenization algorithm and implement it on the filtered corpus. We obtain a vocabulary from the trained tokenizer. We are then able to pretrain a deep language model (\textit{RoBERTa-TR-medium}) using the pretrained tokenizer and obtained vocabulary. We lastly fine-tune our model on several downstream tasks including Sentiment Analysis (SA) and Named Entity Recognition (NER).}
    \label{fig:pipeline}
\end{figure}

We develop a pipeline, illustrated in Figure \ref{fig:pipeline}, which consists of collecting and cleaning the training corpus, training a tokenizer with a fixed-length vocabulary, and lastly pretraining a deep language model by using the selected tokenizer and its vocabulary. We are then able to fine-tune the model on different downstream tasks to evaluate the performance of the tokenizer.

We use the OSCAR deduplicated corpus for pretraining our language model \cite{Suarez:2019, OscarHF:2021}. OSCAR is a multilingual corpus that is obtained by filtering of the Common Crawl corpus that maintain an open repository of publicly available web pages. We use the split of this corpus prepared for Turkish. However, we observe that this split includes many documents in languages other than Turkish. We thereby filter out 95,152 documents that are not in Turkish by using an automated language detector \citep{Shuyo:2010}. The filtering process results in 11,501,370 documents for pretraining.

The tokenization process, depicted inside a dashed rectangle in the figure, is conducted in three steps: (i) Applying normalization to clear the invalid characters from the text, (ii) training the tokenizer according to a predetermined vocabulary size (except Character-level), and (iii) processing the corpus with the trained tokenizer to obtain a tokenized pretraining data. We apply lowercase conversion and NFC normalization\footnote{Unicode normalization is important for Turkish, since there are special characters (ç, ğ, ı, ö, ş, ü) in the Turkish alphabet that are not observed in English. We note that NFC Unicode normalization provides all letters in Turkish.}. 

We pretrain language models using Turkish (TR) text, with RoBERTa pretraining procedure and configuration \cite{Liu:2019}, but smaller in terms of the number of layers, attention heads, and hidden size (we follow the same architecture as BERT-medium \cite{Devlin:2019}). We thereby call the model as \emph{RoBERTa-TR-medium}. We determine the vocabulary size based on the number of parameters of the models. Similar to BERT \cite{Devlin:2019}, the number of parameters associated with the vocabulary constitutes 20\% of the whole model. The vocabulary size for tokenizers are therefore 16.6k tokens, except for Character-level. The calculation of vocabulary size is given as $|V| = (M \times R)/H,$ where $|V|$ is the number of tokens in vocabulary (i.e. the vocabulary size), $M$ is the number of total parameters in the language model, $R$ is the ratio of the vocabulary size to the whole model, and $H$ is the hidden dimension size (in our case, $M$ is approximately 42.7, $H$ is 512).

The pretraining details of our medium model is given in Table \ref{tab:details_of_pretraining}. Since we examine the effect of different tokenization strategies in Turkish, we keep the pretraining procedure computationally simpler because extensive pretraining might overshadow possible advantages of tokenization algorithms. When a model is extensively pretrained, the performance can converge to high scores, even with Character-level encoding \cite{Xue:2021}. Nevertheless, we compare the results of our model with the current state-of-the-art performance for sanity check, i.e. the rationality of our results. To do so, we employ the BERTurk model \cite{Schweter:2020}, which is a Turkish pretrained version of BERT-base \cite{Devlin:2019}. We therefore provide the configuration of BERTurk along with our model's configuration in the table (we did not pretrain BERTurk, but fine-tune it on our downstream tasks).

We use AdamW \cite{Loshchilov:2019} optimizer ($\beta_1$ is 0.90, $\beta_2$ is 0.98, and $\epsilon$ is 1e-6), linear scheduling with a warmup ratio of 1e-2 and peak learning rate of 5e-5, and gradient accumulation with 22 steps. Other hyperparameters are set to the RoBERTa configuration \cite{Liu:2019}. 

\subsection{Fine-tuning Tasks}
\label{sec:fine-tuning-tasks}

The performances of the pretrained models with different tokenizers are evaluated by fine-tuning the models on six downstream tasks. The tasks and datasets used for fine-tuning are explained as follows:

\begin{itemize}[leftmargin=*]

    \item \textbf{News Classification}: Given a set of news articles, this task aims to classify each document into a predetermined set of classes, i.e. the task is text sequence classification. We use Turkish news classification datasets provided by \citet{Toraman:2011}. Merging two datasets from two different news resources results in approximately 7.5k news instances. The news articles are given under eight news topics or categories; namely sports, economy, national, world, politics, columnists, health, and culture-art.
    
    \item \textbf{Hate Speech Detection}: The aim of this task is to determine whether a given text sequence includes hate speech towards other individuals or communities with different backgrounds. Hate Speech Detection is a challenging problem with a limited number of resources in the literature, since there is no decisive consensus on the definitions of the hate or offensive speech, and hate language can have various forms in natural language. In this study, we use a recent hate speech dataset in Turkish, curated by \citet{Toraman:2022}. Data instances are tweets from different hate speech domains including gender, religion, and politics. There are 100k tweets distributed equally among five hate speech domains annotated as hate, offensive, and normal.  
    
    \item \textbf{Sentiment Analysis}: Sentiment Analysis is a task of text sequence classification to find the author's sentimental state. We use a Turkish dataset including movie reviews prepared by \citet{Demirtas:2013}. The reviews are labeled as having positive and negative sentiments. The dataset is balanced and contains approximately 5.3k instances for each sentiment class.
    
\begin{table}[t]
    \centering
    \begin{tabular}{|lr|r|}
    \hline
         & \textbf{BERTurk-base} & \textbf{RoBERTa-TR-medium} \\
    \hline
        Parameters & 110.62 M & 42.69 M \\
        Train data & 35 GB & 27 GB \\
        Layers & 12 & 8 \\
        Heads & 12 & 8 \\
        Hidden size & 768 & 512 \\
        Batch size & n/a & 264 \\
        Max length & 512 tokens & 514 tokens \\
        Train time & 9.63 days & 2 days* \\
        Hardware & TPU v3-8 & 2x Nvidia RTX2080 Ti \\
    \hline
    \end{tabular}
    \caption{Details of pretraining configurations for BERTurk and our model, \emph{RoBERTa-TR-medium}. (*Train time and hardware are given for a vocabulary size of 16.6k tokens. Train time can differ for other vocabulary sizes as we report detailed information in Section \ref{section:discussion}).}
    \label{tab:details_of_pretraining}
\end{table}

    \item \textbf{Named Entity Recognition}: Named Entity Recognition is a token classification task to predict pre-determined named entities in a text sequence, such as person and location. We use the benchmark dataset \cite{Tur:2003} including Turkish news articles. The dataset contains approximately 32.5k sentences and three named entity classes given as person, location, and organization. The named entities in the dataset are annotated with the IOB2 \cite{Ramshaw:1995} tags, such that each entity chunk starts with B-$<$class$>$, and continues with I-$<$class$>$, e.g. New York has the tags of B-<LOCATION> and I-<LOCATION>.
    
    \item \textbf{Semantic Text Similarity}: Semantic similarity between two text sequences is measured in this task. The sentence pairs are annotated in a scale between 0 (i.e. no semantic similarity) and 5 (i.e. semantically equivalent) according to their similarity degree. Different from the classification tasks, STS is handled as a regression problem. To evaluate the performance of the model, the correlation between the ground truth and model predictions is taken into consideration. We use a Turkish STS dataset that is the translation of the STSb dataset \citep{Beken:2021}. The domain of the sentence pairs vary from news articles to online forum messages. The dataset includes approximately 8.6k sentence pairs in total.
    
    \item \textbf{Natural Language Inference}: Given two sentences, the aim of Natural Language Inference is to predict whether the latter is inferred by the former. The dataset includes three types of semantic relations: The first sentence can entail the second one (\textit{entailment}), the sentences can be irrelevant to each other (\textit{neutral}), or the first sentence can contradict the second one (\textit{contradiction}). We use a Turkish NLI dataset which is the translated version of the SNLI dataset \cite{Budur:2020}. The dataset includes approximately 570k sentence pairs.

\end{itemize}

\begin{table}[t]
    \centering
    \setlength{\tabcolsep}{2pt}
    \begin{tabular}{|ll|c|c|c|c|c|c|}
    \hline
            & & \textbf{News} &
            \textbf{Hate Speech} &
            \textbf{Sentiment} &  \textbf{Named Entity} & \textbf{Semantic Text} & \textbf{Natural Language} \\
        & & \textbf{Classification} &
        \textbf{Detection} &
        \textbf{Analysis} &  \textbf{Recognition} & \textbf{Similarity} & \textbf{Inference} \\
     \hline
        \multirow{3}{*}{\rotatebox[origin=c]{90}{R-TR-m}} & Epochs & 10 & 5 & 10 & 10 & 10 & 3 \\
        & Max. length & 514 & 256 & 514 & 514 & 514 & 514 \\
        & Batch size & 32 & 32 & 32 & 16 & 16 & 16 \\
        \hline
        \multirow{3}{*}{\rotatebox[origin=c]{90}{BERT}} & Epochs & 3 & 3 & 3 & 3 & 25 & 3 \\
        & Max. length & 256 & 256 & 256 & 256 & 256 & 256 \\
        & Batch size & 32 & 32 & 32 & 16 & 32 & 16 \\
        \hline
        & Learning rate & 1e-5 & 1e-5 & 1e-5 & 1e-5 & 1e-5 & 1e-5 \\
        & Train Size & 6,786 & 90,000 & 9,595 & 29,332 & 7,765 & 512,130 \\
        & Test Size & 754 & 10,000 & 1,067 & 3,259 & 863 & 56,903 \\
    \hline
    \end{tabular}
    \caption{Details of fine-tuning configurations for different downstream tasks. \emph{R-TR-m} refers to our model, RoBERTa-Turkish-medium, and \emph{BERT} refers to BERTurk. The task configurations can change due to space complexity and data size. We apply constant learning rate for all tasks, except linear decay learning rate in NLI. When Character-level tokenizer is used, we set the number of epochs to 10, max. sequence length to 1024, and batch size to 12, for all tasks.}
    \label{tab:fine-tune_config}
\end{table}

\section{Experiments}
\label{section:experiments}
We conduct two experiments in this study. First, we compare the performances of tokenization methods for Turkish downstream tasks. Second, we analyze the effect of vocabulary size on the downstream task performance. For each experiment, we describe our experimental design, and then report the results.

\subsection{Comparison of Tokenization Methods}

\subsubsection{Experimental Design}
We compare the performance of tokenizers using \emph{RoBERTa-TR-medium} on Turkish downstream tasks (\textbf{RQ-1}). We use a fixed size of vocabulary, 16.6k tokens, in this experiment to understand how different tokenizers would perform under the same configuration. Furthermore, we analyze the performance of our medium-size model in comparison with a larger state-of-the-art model. To do so, we report the performance of BERTurk \cite{Schweter:2020}, a Turkish model with a size similar to BERT-base \cite{Devlin:2019}.

We fine-tune the models, which are pretrained using different tokenizers, on the downstream tasks given in Section \ref{sec:fine-tuning-tasks}. For fine-tuning our models, the configurations and hyperparameters along with dataset sizes are given in Table \ref{tab:fine-tune_config}. We measure weighted precision, recall, and F1 score for all tasks, except STS where Pearson correlation is reported with p-value. We apply 10-fold cross-validation and report the average scores. We determine statistically significant differences between the means, which follow non-normal distributions, by using the two-sided Mann-Whitney U (MWU) test at \%95 interval with Bonferroni correction. 

\subsubsection{Experimental Results}
We report the fine-tuning results in Table \ref{tab:finetuning}. Our main observations and answers to \textbf{RQ-1} can be summarized as follows.

\begin{table*}[t]
\setlength{\tabcolsep}{1.1pt}
\small
\centering
\begin{tabular}{|llccc|ccc|ccc|ccc|cc|ccc|}
\hline
& & \multicolumn{3}{c|}{\textbf{News}} & \multicolumn{3}{c|}{\textbf{Hate Speech}} & \multicolumn{3}{c|}{\textbf{Sentiment}} & \multicolumn{3}{c|}{\textbf{Named Entity}} & \multicolumn{2}{c|}{\textbf{Semantic Text}} & \multicolumn{3}{c|}{\textbf{Natural Language}} \\
& & \multicolumn{3}{c|}{\textbf{Classification}} & \multicolumn{3}{c|}{\textbf{Detection}} & \multicolumn{3}{c|}{\textbf{Analysis}} & \multicolumn{3}{c|}{\textbf{Recognition}} & \multicolumn{2}{c|}{\textbf{Similarity}} & \multicolumn{3}{c|}{\textbf{Inference}} \\

& & \textbf{P} & \textbf{R} & \textbf{F1} & \textbf{P} & \textbf{R} & \textbf{F1} & \textbf{P} & \textbf{R} & \textbf{F1} & \textbf{P} & \textbf{R} & \textbf{F1} & \textbf{corr} & \textbf{p-value}
& \textbf{P} & \textbf{R} & \textbf{F1} \\

\hline
& BERT & 0.918 & 0.917 & 0.917 & 0.781 & 0.781 & 0.781 & 0.927 & 0.927 & 0.927 & 0.935 & 0.955 & 0.945 & 0.862 & <1e-178 & 0.852 & 0.852 & 0.852 \\

\hline

\multirow{5}{*}{\rotatebox[origin=c]{90}{R-TR-medium}} & Char & 0.715 & 0.723 & 0.713 & 0.606 & 0.609 & 0.607 & 0.812 & 0.812 & 0.812 & 0.730 & 0.788 & 0.757 & 0.256 & <1e-4 & 0.620 & 0.619 & 0.619 \\

& BPE & \textbf{0.886} & \textbf{0.885} & \textbf{0.885} $\bullet$ & 0.742 & 0.737 & 0.738 & 0.882 & 0.881 & 0.881 $\circ$ & 0.851 & 0.883 & 0.866 $\circ$ & 0.487 & <2e-32 & 0.772 & 0.772 & 0.772 \\

& WP & 0.882 & 0.881 & 0.881 $\circ$ & \textbf{0.745} & \textbf{0.745} & \textbf{0.745} $\bullet$ & \textbf{0.884} & \textbf{0.884} & \textbf{0.884} $\bullet$ & \textbf{0.858} & \textbf{0.893} & \textbf{0.875} $\bullet$ & \textbf{0.718} & <3e-92 $\bullet$ & \textbf{0.778} & \textbf{0.778} & \textbf{0.778} $\bullet$ \\

& Morph & 0.869 & 0.868 & 0.867 & 0.726 & 0.727 & 0.726 & 0.824 & 0.823 & 0.823 & 0.839 & 0.872 & 0.855 & 0.655 & <5e-63 $\circ$ & 0.768 & 0.768 & 0.768 \\ 

& Word & 0.857 & 0.857 & 0.856 & 0.647 & 0.649 & 0.648 & 0.805 & 0.805 & 0.805 & 0.791 & 0.740 & 0.764 & 0.492 & <2e-16 & 0.603 & 0.598 & 0.595 \\

\hline
\end{tabular}
\caption{Fine-tuning results of different tokenizers (rows) on six downstream tasks (columns) using Turkish datasets. The average of 10-fold cross-validation is reported in terms of weighted precision (P), recall (R), and F1 score. \emph{BERT} refers to BERTurk, which is structurally similar to BERT-base, but pretrained for Turkish text. For STS, Pearson correlation (corr) is reported with p-value. \emph{R-TR-medium} refers to our pretrained model for Turkish text, \emph{RoBERTa-Turkish-medium}, along with each tokenization method. \emph{Char} refers to Character-level tokenizer, \emph{BPE} refers to Byte Pair Encoding, \emph{WP} refers to WordPiece, \emph{Morph} refers to Morphological-level tokenizer, and \emph{Word} refers to World-level tokenizer. The highest score among tokenizers for each task is given in bold. The symbol ``$\bullet$" indicates statistical significant difference at a 95\% interval (with Bonferroni correction $p<0.0125$) in pairwise comparisons between the highest performing method and others (except the ones with ``$\circ$").}

\label{tab:finetuning}
\end{table*}

\textbf{WordPiece and BPE are highest performing tokenizers in Turkish language modeling}.
WordPiece and BPE are de facto standard tokenizers that are practically employed in language modeling. Indeed, we find that WordPiece statistically significantly outperforms other tokenizers in most of the tasks in our experiments for Turkish. The only exceptions are that BPE has higher scores in News Classification, but the difference between the performances of BPE and WordPiece is not statistically significant in that case. Similarly, WordPiece has higher score than BPE in Sentiment Analysis and Named Entity Recognition, but the differences are not statistically significant as well. We thereby argue that WordPiece and BPE are highest performing tokenizers in Turkish, yet WordPiece has better performance than BPE in the majority of tasks.

\textbf{Word-level tokenizer performs poorly due to many unknown tokens}. Word-level tokenizer performs poorly compared to BPE, WordPiece, and Morphological-level tokenizers, possibly due to poor utilization of the available vocabulary capacity. We therefore examine the ratio of the unknown tokens to all tokens in the fine-tuning datasets, and report them in Table \ref{tab:unk_tokens}. It is apparent that almost half the tokens are unknown to the model when Word-level tokenizer is used in all tasks. The reason why Word-level still achieves comparable results with other tokenizers might be that the model is trained with the Masked Language Modeling task and gains an ability to infer meaning even with many unknown tokens.

\textbf{Morphological-level has competitive results with state-of-the-art tokenizers}. The differences between the weighted F1 scores of Morphological-level tokenizer and the best performing tokenizer is statistically significant but very small (between 0.01 and 0.02) in all tasks, except that it is approximately 0.06 in Sentiment Analysis and Semantic Text Similarity (STS). However, this difference is not statistically significant in STS. We thereby argue that Morphological-level tokenizer achieves competitive results with de facto tokenizers. Moreover, the performance of Morphological-level tokenizer is always better than those of Character-level and Word-level tokenizers. We argue that suffixes can provide useful information for language modeling in Turkish. The poor performance compared to de facto tokenizers can be attributed to two observations. First, the method has dependency on the performance of the morphological analyzer, Zemberek \cite{Akin:2007}, which we employ for obtaining prefixes and suffixes. We observe possible errors such as wrong morphemes in the output of morphological analyzer as reported Table \ref{tab:zemberek_output}. For instance, the word "İstanbullular" (translated to "People of Istanbul") is not tokenized correctly; however, the word contains inherit information, such as hometown (\#\#lu) and plural (\#\#lar). Second, Morphological-level tokenizer is limited in terms of word stems since roots are not split into smaller pieces in this approach, increasing the likelihood of observing unknown tokens, as observed in Table \ref{tab:unk_tokens}.

\textbf{Character-level tokenizer has no significant benefit}.
Character-level tokenizer achieves the worst performance for Turkish in most tasks. The reason could be that our medium models might be inadequate to comprehend the relations among characters, which could be better modeled by larger language models \cite{Xue:2021}. However, we argue that the size and architecture of larger models could be inefficient to outperform de facto tokenizers, as we address in Section \ref{section:tokenization_methods}.

\begin{table*}[t]
\setlength{\tabcolsep}{1.1pt}
\centering
\begin{tabular}{|l|c|c|c|c|c|c|}
\hline
& \textbf{News} & \textbf{Hate Speech} & \textbf{Sentiment} & \textbf{Named Entity} & \textbf{Semantic Text} & \textbf{Natural Language} \\
& \textbf{Classification} & \textbf{Detection} & \textbf{Analysis} & \textbf{Recognition} & \textbf{Similarity} & \textbf{Inference} \\
\hline
BPE & 0.000 & 0.000 & 0.000 & 0.000 & 0.000 & 0.000 \\
WordPiece & 1.447e-6 & 4.292e-6 & 4.824e-6 & 0.000 & 0.000 & 0.000 \\
Morph-level & 0.021 & 0.171 & 0.183 & 0.018 & 0.024 & 0.008 \\ 
Word-level & 0.587 & 0.515 & 0.502 & 0.457 & 0.522 & 0.457 \\
\hline
\end{tabular}
\caption{Ratios of unknown tokens to all tokens in the fine-tuning datasets that are used in Table \ref{tab:finetuning}.}

\label{tab:unk_tokens}
\end{table*}
\begin{table*}[t]
\setlength{\tabcolsep}{1.1pt}
\centering
\begin{tabular}{|l|cccc|}
\hline
\textbf{} &  & & \textbf{Output} & \\
\hline
Sample Sentence & İstanbullular & güneşin & tadını & çıkarabildiler\\
True Parsing & İstanbul \#\#lu \#\#lar & güneş \#\#in & tat \#\#ı \#\#nı & çık \#\#ar \#\#abil \#\#di \#\#ler\\
Zemberek & İstanbullular & güneş \#\#in & tad \#\#ın \#\#ı & çıkar \#\#abil \#\#di \#\#ler\\
\hline
\end{tabular}
\caption{Tokenization output of the true parsing, and a morphological analysis tool, Zemberek, considering Turkish syntax rules. Sample sentence is "İstanbullular güneşin tadını çıkarabildiler" (translated as "People of Istanbul were able to enjoy the sun").}
\label{tab:zemberek_output}
\end{table*}

\textbf{Medium models can be competitive to larger ones}. We expect that the performance of our medium models is lower than larger models, i.e. BERTurk, due to the computational advantages of larger models. However, we find that the performance gap is narrow for particular tasks. Our 3-times smaller model recovers 97\% of BERTurk's performance in News Classification, 95\% in Hate Speech, 95\% in Sentiment Analysis, 93\% in Named Entity Recognition, 83\% in Semantic Textual Similarity, and 91\% in Natural Language Inference. One possible reason of the relatively lower recovery score of STS task is that it is a regression task. In classification, output logits are mapped to classes. Since there is no such quantization in regression, there can be more deviations in the correlation between ground truths and predictions. 

\subsection{Analysis of Vocabulary Size} 

\subsubsection{Experimental Design}
In the previous experiments, we fix the vocabulary size for all tokenizers except Character-level tokenizer. However, vocabulary embeddings contribute to the total number of model parameters and the effect of the vocabulary size can vary among different tokenizers (\textbf{RQ-2}). We thereby design an experiment that measures the performances of tokenizers with changing vocabulary sizes. We note that in the ultimate case when the vocabulary size tends to infinite, every possible character combination in corpus is assigned a representation in the vocabulary. In such a case, the modeling becomes similar to conventional word embeddings, such as word2vec \citep{Mikolov:2013}. On the other extreme, the need for contextual representation of a given token increases as vocabulary size gets smaller. In other words, a single token is expected to reflect a wide variety of contextual meanings due to limited vocabulary size. 

In this experiment, we fix the hyperparameters of the Transformer blocks in the architecture, e.g. the number of layers and hidden size, and adjust the vocabulary size such that the number of parameters attributed to the vocabulary constitutes 10, 20, 30, 40, and 50 percent of the entire model. Since Character-level tokenizer has a fixed vocabulary size, we exclude it from this experiment. 

This analysis requires to train a separate language model for five vocabulary sizes and four tokenization methods, resulting in a total number of 20 models. Considering six downstream tasks, we would have 120 experimental runs. We therefore decide to select two important tasks among six tasks, for the sake of efficiency and carbon footprint. We select a text classification task, Sentiment Analysis, and a token classification task, Named Entity Recognition. In fine-tuning, we apply 10-fold cross-validation and report the average of weighted F1 scores.

\begin{figure}
    \centering
    \includegraphics[width=\columnwidth]{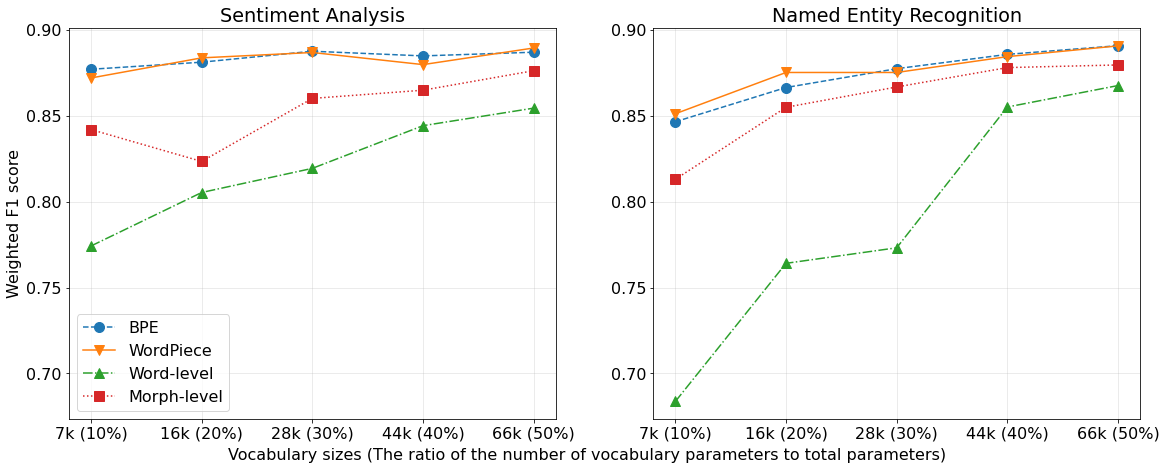}
    \caption{The effect of varying the vocabulary size for different tokenization methods. The results are reported for a text classification task, Sentiment Analysis, and a token classification task, Named Entity Recognition.}
    \label{fig:vocab_size}
\end{figure}

\subsubsection{Experimental Results}
The results of varying vocabulary size for different tokenizers are given in Figure \ref{fig:vocab_size}. We report the vocabulary sizes by removing fractional part of decimal numbers (e.g. 16.6k is given as 16k). Our main observations and answers to \textbf{RQ-2} are listed as follows:

\textbf{We observe an increasing pattern for the performance of all tokenizers as the vocabulary size increases}. Increasing vocabulary size can result in less number of unknown tokens for Morphological-level and Word-level tokenizers. BPE and WordPiece can benefit from higher level tokens by merging subword tokens when vocabulary size increases, e.g. \textit{New Y \#\#or \#\#k} can be represented as \textit{New York} when vocabulary size is sufficient. In terms of downstream tasks, Named Entity Recognition can benefit from increasing vocabulary size more than Sentiment Analysis, since Named Entity Recognition is a token classification task that depends on the performance of predicting individual tokens. The reason of smaller improvements in Sentiment Analysis could be that it is a sequence classification task that can tolerate individual unknown tokens while predicting sentiment of the given sequence.

\textbf{Performance improvement gets faster saturation for de facto tokenizers compared to others}. De facto tokenizers, i.e. BPE and WordPiece, do not dramatically benefit from larger vocabulary sizes, which indicates a saturation in performance at smaller sizes. A potential cause for this saturation might be that BPE and WordPiece can tokenize almost any input in the fine-tuning datasets with a vocabulary size greater than 16k (i.e. the ratios of unknown tokens to all tokens are mostly zeros in Table \ref{tab:unk_tokens}). Furthermore, they outperform Morphological-level and Word-level tokenizers in both tasks. However, the performance gap diminishes as the vocabulary size increases, probably due to less unknown tokens. This also reduces the impact of tokenization on the model performance. In terms of downstream tasks, while Sentiment Analysis has performance saturation in small number of vocabulary sizes (e.g. 7k and 16k), Named Entity Recognition gets saturation after 16k probably due to the fact that it is a token classification task.

\textbf{Very large vocabulary sizes have less practical advantage, specifically for de facto tokenizers}. When the vocabulary size exceeds 66k tokens, the performance improvement may become infeasible due to the increasing computational complexity. Instead, the computational resources can be dedicated to increase other resources, such as the number of parameters in the Transformer blocks. For de facto tokenizers, this situation is more apparent because their performances are already saturated at even smaller vocabulary sizes. We observe that the ratio of the number of vocabulary parameters to the total number of model parameters can be empirically chosen as 20\% for de facto tokenizers and 40\% for others, in order to satisfy the trade-off between model size and performance.

\section{Broader Impact and Ethical Concerns}
\label{section:discussion}

Since we pretrain large language models and fine-tune them on several downstream tasks for experimental evaluations, we would like to emphasize broader impact and ethical concerns \cite{Mitchell:2019, Bender:2021, Baeza-Yates:2022}. We thereby provide a discussion on our study's broader impact, transparency, responsibility, and fairness in this section.

\subsection{Broader Impact}
We anticipate that our study would have a broader impact on the research community that focuses on the impact of tokenization and vocabulary size on language modeling. Although the models are medium-sized, we observe that their downstream performances are comparable with base models (see Table \ref{tab:finetuning}). Our research elaborates on an agglutinative low-resource language, Turkish, and our findings can provide guidance to other researchers that study similar languages. We also provide performance measurements for a diverse set of downstream tasks which can be useful for various real-life applications, such as recommendation systems and finance.

\subsection{Transparency}
In order to provide a transparent modeling \cite{Mitchell:2019}, we explain all details regarding text corpus used in pretraining our language models in Section \ref{sec:our_model}, as well as the details of the algorithms used throughout obtaining the language models in Section \ref{section:tokenization_methods} and model configurations in Section \ref{sec:our_model}. The details of how we conduct the experimental evaluations are also reported in Section \ref{sec:fine-tuning-tasks} and Section \ref{section:experiments}. 

\subsection{Responsibility}
There is an increasing environmental awareness among the Machine Learning community about responsible training such as the carbon footprint of extensive model training \cite{Henderson:2020, Bender:2021}. We estimate the carbon footprint of our study based on the energy usage of GPUs for pretraining and fine-tuning, and report them in Table \ref{tab:pretrain_cost} and Table \ref{tab:fine-tune_cost}, respectively. We report execution time in hours and electrical energy consumption in kWh. It is assumed that the power consumption during training is equal to the maximum power drain of GPUs and they operate at maximum power utilization (250W). This estimation ignores the carbon footprint of CPU utilization and the manufacturing costs of the hardware. We note that the emissions for different tokenization methods are close to each other, and report the cost for a single tokenization method, WordPiece, in this analysis. 

\begin{figure}
    \centering
    \includegraphics[width=0.9\columnwidth]{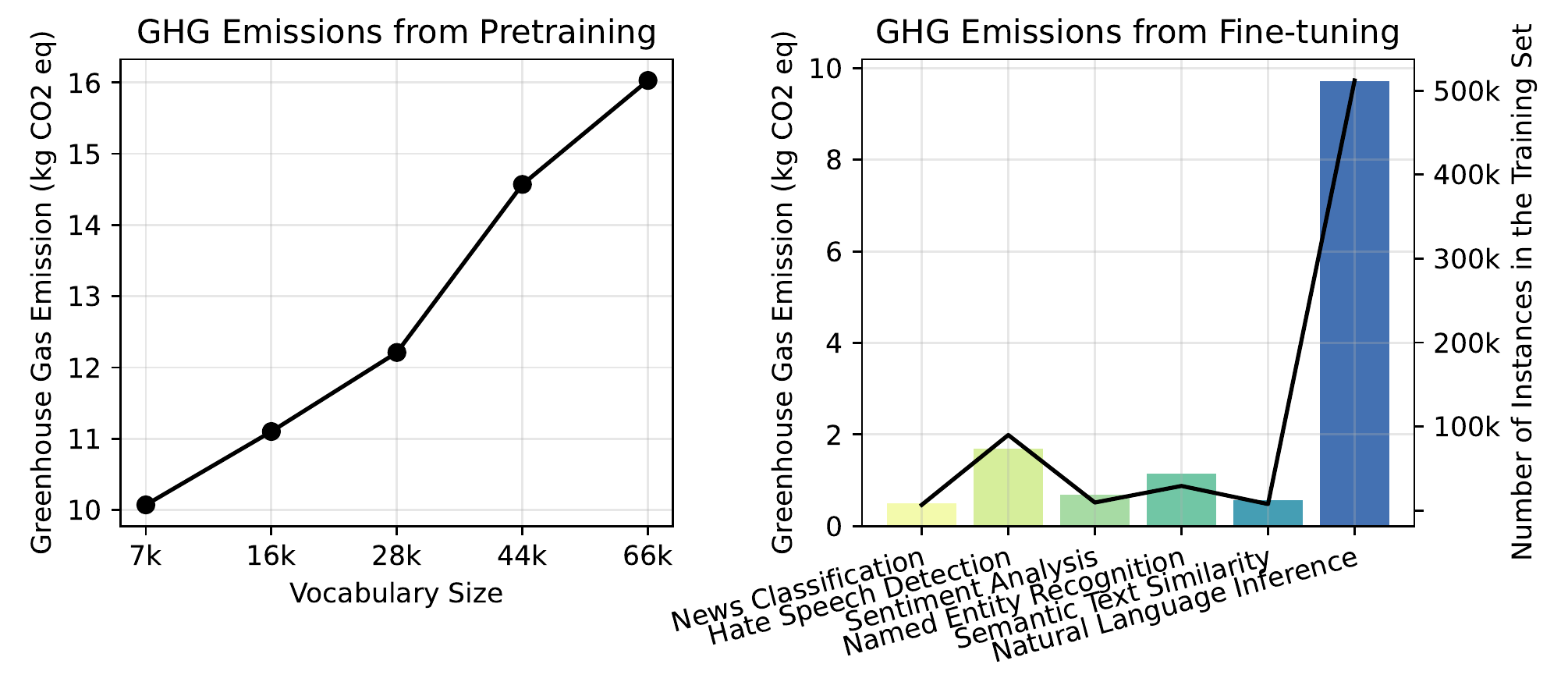}
    \caption{Carbon footprint (in terms of kg CO${_2}$) of pretraining with a single tokenization method for different vocabulary sizes (left) and fine-tuning for different downstream tasks (right). The numbers of training instances used in fine-tuning are represented by a line at the right subplot with their corresponding values at the second y-axis.}
    \label{fig:ghg_pretraining}
\end{figure}

Based on the calculations of execution time and energy consumption, we estimate the carbon footprint of our models in terms of greenhouse gas (GHG) emission in kg CO${_2eq}$. We then plot the carbon footprint of pretraining with different vocabulary sizes and fine-tuning on different downstream tasks in Figure \ref{fig:ghg_pretraining}. We note that for the GHG emission of pretraining, the vocabulary sizes given at the x-axis have an exponential scale, starting from 7k to 66k. We observe a linearly increasing trend in GHG emissions as the vocabulary size increases, which indicates an exponential growth in environmental damage. We underline that the performance gain in Figure \ref{fig:vocab_size} comes at a cost of increasing environmental damage, and therefore suggest that a reasonably smaller vocabulary size is a preferable choice for pretraining. We further observe that the greenhouse gas emissions caused by the fine-tuning experiments are in correlation with the size of the utilized training set. 

For the conversion of electrical energy usage to CO${_2eq}$ Greenhouse Gas (GHG) emission, we use a local conversion factor specified by the Turkish government \citep{Ministry:2020}. The specified value is an upper bound of the spontaneous value reported by electricitymap \citep{Electricitymap:2022}. Based on our estimation for the GHG release, a lower bound for the social carbon cost (SCC) of the experiments in our study can be approximated as \$117.42 with a value of \$300 per ton of CO${_2}$ \cite{Kikstra:2021}. 

\begin{table}[t]
    \centering
    \setlength{\tabcolsep}{4pt}
    \begin{tabular}{|l|c|c|c|c|c|}
    \hline
    & \multicolumn{5}{c|}{\textbf{Vocabulary Size}} \\
    \cline{2-6}
        & \textbf{7k} & \textbf{16k} & \textbf{28k} & \textbf{44k} & \textbf{66k}  \\
    \hline
    \cline{1-6}
        GPU Hours (h) & 2 $\times$ 36.3h & 2 $\times$ 40h & 2 $\times$ 44h & 2 $\times$ 52.5h & 2 $\times$ 57.75h  \\
        Energy Consumption (kWh) & 18.15 & 20.00 & 22.00 & 26.25 & 28.875  \\
    \hline
    \end{tabular}
    \caption{Energy consumption for pretraining with a single tokenization method for different vocabulary sizes.}
\label{tab:pretrain_cost}
\end{table}

\begin{table}[t]
    \centering
    \setlength{\tabcolsep}{1.5pt}
    \begin{tabular}{|l|c|c|c|c|c|c|}
    \hline
        & \textbf{News} & \textbf{Hate Speech} & \textbf{Sentiment} & \textbf{Named Entity} & \textbf{Semantic Text} & \textbf{Natural Language}  \\
        & \textbf{Classification} & \textbf{Detection} & \textbf{Analysis} & \textbf{Recognition} & \textbf{Similarity} & \textbf{Inference}  \\
    \hline
    \cline{1-7}
        GPU Hours (h) & 2 $\times$ 1.77h & 2 $\times$ 6.08h & 2 $\times$ 2.50h & 1 $\times$ 8.33h & 2 $\times$ 2.05h & 2 $\times$ 35.00h \\
        Energy Consump. (kWh) & 0.89 & 3.04 & 1.25 & 2.08 & 1.03 & 17.5 \\
    \hline
    \end{tabular}
    \caption{Energy consumption for fine-tuning with a single tokenization method for different downstream tasks.}
\label{tab:fine-tune_cost}
\end{table}

\subsection{Fairness}
The pretraining text corpus in our study, i.e. the OSCAR corpus, has millions of web page texts, a part of which may include biased contents ignoring the fairness principle towards all communities and individuals with a variety of backgrounds and profiles. Moreover, the fairness of the tokenization algorithms and Transformer-based language models is still in debate \cite{Baeza-Yates:2022}. We are not able to make a judgment on the fairness of the corpus and model; since, to the best of our knowledge, there is no available automatic tool to assess fairness. Nevertheless, we acknowledge that the researchers and practitioners should provide fair algorithms, data, and models. The bias can be removed by filtering the pretraining corpus or analyzing the fairness of algorithms used in this study, however we leave the study of such filtering and analysis to future work, since it would require a dedicated effort to develop such algorithms, but the scope of this study is to compare empirical performance of different tokenizers and vocabulary sizes.

\section{Conclusion}
\label{section:conclusion}
We provide a comprehensive study that examines the impact of tokenization in Turkish, which is a low-resource language with a limited number of pretrained deep language models. In order to accomplish this task, we pretrain a medium-sized language model, called \emph{RoBERTa-TR-medium}, with different tokenization algorithms and varying vocabulary sizes. Our language models are publicly available, so that other researchers and practitioners can benefit from our models. This would provide less electrical energy and memory usage with better carbon footprint in return.

Our experimental results, supported by statistical tests, can shed light on the role of tokenization in language modeling, specifically for morphologically rich languages. We find that Morphological-level tokenizer is competitive with de facto tokenizers, i.e. BPE and WordPiece. Our models, which are approximately 3-times smaller than state-of-the-art larger models, can recover 97\% of the performance of the larger one. We also show that increasing the vocabulary size improves the performance of Morphological and Word-level tokenizers more than that of de facto tokenizers. We suggest that the ratio of the number of vocabulary parameters to the total number of model parameters can be empirically chosen as 20\% for de facto tokenizers and 40\% for others, for the trade-off between model size and performance. 

In future work, we plan to extend our experiments to other agglutinative languages, such as Finnish and Hungarian, and other tokenization algorithms such as SentencePiece \cite{Kudo:2018a}. Morphological disambiguation \cite{Hakkani:2018} can be used to improve the quality of morphological analysis, yielding to potential improvements in Morphological-level tokenization. We also plan to focus more on AI ethics for the impact of tokenization for pretraining language models, including but not limited to filtering bias in pretraining text corpora and analysis of tokenization algorithms in terms of fairness. 

\bibliography{tokenization}
\bibliographystyle{ACM-Reference-Format}

\end{document}